\documentclass[letterpaper, 10 pt, conference]{ieeeconf}  

\IEEEoverridecommandlockouts                              

\usepackage{amsmath,amsfonts}
\usepackage{algorithmic}
\usepackage{algorithm}
\usepackage{array}
\usepackage[caption=false,font=normalsize,labelfont=sf,textfont=sf]{subfig}
\usepackage{textcomp}
\usepackage{stfloats}
\usepackage{url}
\usepackage{verbatim}
\usepackage{graphicx}
\usepackage{cite}
\usepackage{amssymb}
\usepackage{dsfont}
\usepackage{booktabs}

\newcommand{\eshift}{\epsilon_{\text{shift}}}
\newcommand{\epre}{\epsilon_{\text{pre}}}
\newcommand{\ind}{\mathds{1}}

\title{\LARGE \bf
Beyond Domain Randomization: Real-World Policy Adaptation \\
in Learned Dynamics for Legged Robots
}

\author{Rui Dai$^{1}$, Shunpeng Yang$^{2}$, Zhen Fu$^{2, 3}$, Maolin Lei$^{1}$, Jiaqi Song$^{3}$, Hua Chen$^{2}$, and Nikos Tsagarakis$^{1}$
\thanks{$^{1}$Humanoids and Human-Centered
Mechatronics Research Line, Istituto Italiano di Tecnologia (IIT), 16163 Genova, Italy.}%
\thanks{$^{2}$LimX Dynamics, 518055 Shenzhen, China.}%
\thanks{$^{3}$School of Automation and Intelligent Manufacturing (AIM), Southern University of Science and Technology, 518055 Shenzhen, China.}%
}

\begin{document}

\maketitle
\thispagestyle{empty}
\pagestyle{empty}

\begin{abstract}

Reinforcement learning policies are often trained with static domain
randomization for real-world deployment. However, real-world environments often exhibit unmodeled dynamic variations, such as payload changes or friction shifts, 
that lie outside the randomization distribution, leading to performance
degradation upon deployment.
We propose a framework that moves beyond static domain randomization by adapting
a learned dynamics model to the deployed target conditions, with a
concurrent teacher-student paradigm applied to both the policy and dynamics
model.
This framework integrates the dynamics model into an on-policy reinforcement learning
pipeline, letting the performance of the policy align with the empirical
target dynamics. Experimental validations on bipedal locomotion tasks demonstrate
competitive adaptability under extreme physical variations,
including a payload reaching 80\% of self-weight and an oil-lubricated
low-friction surface.

\end{abstract}

\section{Introduction}
\begin{figure*}[thb]
    \centering
    \includegraphics[width=0.99\textwidth]{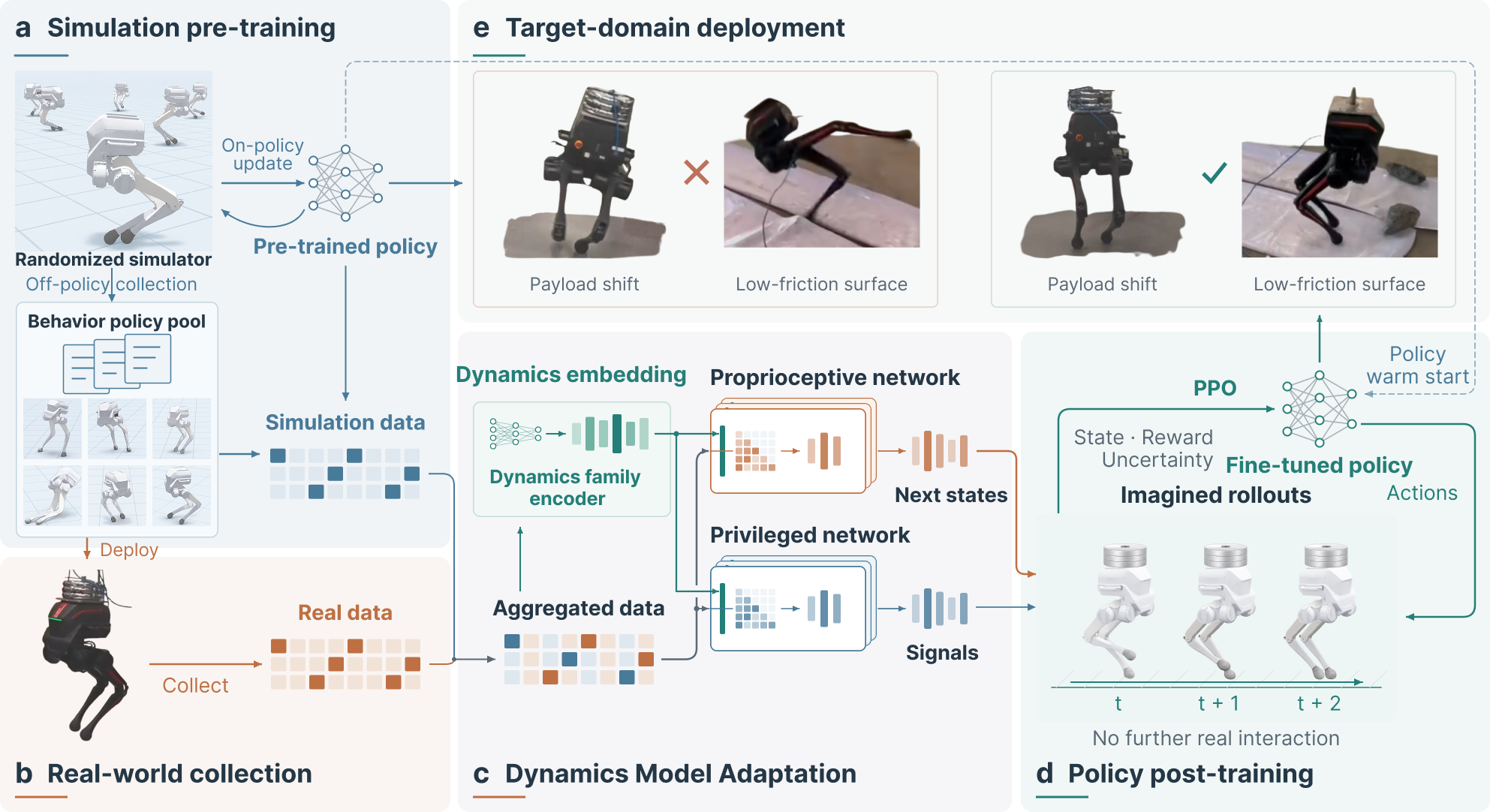}
    \caption{Overview of the framework. The pipeline operates in four phases.
    (a) Simulator online pre-training, where a nominal policy is trained via
    a teacher-student architecture, while a dynamics model is learned
    concurrently using aggregated off-policy data. (b) Real world data collection
    with the pre-trained behavior policies. (c) The aggregated simulation and
    real data are used to fine-tune the dynamics model. (d) The dynamics model
    creates a high-fidelity surrogate to fine-tune the policy against physical variations.}
    \label{fig:framework}
\end{figure*}

Reinforcement learning has recently achieved remarkable success in robotic
control. This success is largely driven by high-fidelity simulators that
enable massive trajectory rollouts and systematic policy
evaluation~\cite{rudin2022learning,hwangbo2019learning}. Within this paradigm,
domain randomization (DR) has established itself as a cornerstone technique to
bridge the reality gap, effectively handling modeling errors by exposing the
agent to a spectrum of physical parameters during
training~\cite{tobin2017domain,peng2018sim}.
However, relying on a static
randomization range during training imposes inherent limitations when facing
significant dynamic variations. Fixed randomization ranges cannot fully capture extreme conditions encountered during deployment, such as substantial payload shifts or severe hardware degradation. Consequently, learned policies can either become overly conservative, sacrificing task performance to maintain stability, or fail catastrophically when faced with large dynamics mismatches~\cite{mehta2020active,kumar2021rma}.

Alternative approaches focus on reducing the
reality gap through system identification (SysID)~\cite{bjelonic2025towards}
or learned actuator networks~\cite{hwangbo2019learning}. While SysID
effectively leverages analytical physics priors to calibrate simulation
parameters, its performance is fundamentally bounded by the structural
assumptions of rigid-body dynamics. Variations that fall outside this
parametric family introduce structural bias,
and
additional interaction data cannot expand the expressive capacity of
SysID~\cite{yu2017preparing}. Learned actuator networks
introduce
neural flexibility to capture nonlinearities, but typically require dense
supervision from high-accuracy torque sensors~\cite{hwangbo2019learning}.
Moreover, as component-level models, they often fail to generalize to
system-level perturbations. Consequently, effective adaptation should benefit from flexible system-level dynamics models that capture effects beyond predefined physical parameterizations and can be learned directly from interaction data without requiring sophisticated sensing setups.

In this work, we incorporate learned system-level dynamics models into policy optimization within a model-based reinforcement learning (MBRL) paradigm tailored to real-world domain adaptation. 
Existing MBRL approaches include methods that augment off-policy training with model-generated transitions~\cite{janner2019trust,yu2020mopo}, as well as methods that combine model and policy learning with continued environment interaction~\cite{hafner2025mastering,wu2023daydreamer}. We instead consider a practical setting in which only a limited batch of real-world interaction data is available from the target domain. 
In the proposed framework, as illustrated in Fig.~\ref{fig:framework}, offline real-world trajectories are collected with diverse behavior policies from simulation pre-training. These trajectories serve solely to align
the dynamics model with the target domain, after which the pre-trained
policy is fine-tuned via on-policy proximal policy optimization (PPO) entirely
inside this surrogate without any further environment interaction in the real
world. This decoupling of offline data collection from on-policy optimization
combines two properties usually held in tension by prior MBRL methods. The dynamics
model is policy-agnostic and ingests data from arbitrary, diverse behavioral
policies, while policy improvement remains on-policy and benefits from the
stability and data-scaling of PPO that have proven crucial for
locomotion~\cite{rudin2022learning}.
We demonstrate the approach on a real robot locomotion task under extreme domain shifts.
Our main contributions are as follows.
\begin{itemize}
    \item We propose a target-specific domain adaptation framework that uses
    real-world data to drive on-policy optimization through a learned dynamics
    model,
    enabling the policy to adapt to real-world physical variations beyond the
    training randomization range.
    \item We design a unified architecture for concurrent pre-training
    of the policy and dynamics model in a single stage. This architecture enables
    both components to operate on proprioceptive
    inputs, facilitating subsequent post-training using the limited observations
    available in the real world.
    \item The framework is empirically validated on a real bipedal robot
    platform, demonstrating competitive adaptability under extreme physical
    variations compared with conventional baselines.
\end{itemize}

\section{Related Work}
\label{sec:related_work}

\noindent\textbf{History-Based Adaptation.} To overcome the limitations of static domain randomization~\cite{tobin2017domain},
recent research has focused on adaptive control strategies through implicit
system identification. Rapid motor adaptation (RMA)~\cite{kumar2021rma} and its
variants~\cite{fu2021minimizing} utilize a history encoder to compress
proprioceptive observations into a latent extrinsic vector. This latent
representation essentially acts as a proxy for system identification, guiding
the policy to react to environmental changes. More recently, Transformer-based
architectures have been employed to perform in-context learning for locomotion,
treating adaptation as a sequence modeling problem~\cite{liu2025locoformer}.
However, these approaches remain fundamentally model-free. They rely on learning
implicit correlations between history and privileged states without
establishing a causal, predictive understanding of the system dynamics.
Consequently, when facing severe disturbances, these implicit encoders struggle
to reconstruct the underlying physical information, leading to reactive rather
than anticipatory adaptation. In contrast, our framework leverages a forward
predictive objective, forcing the policy to optimize against target
dynamics that explicitly respects the causality of physical interactions.

\noindent\textbf{Learned Dynamics.} To bridge analytical simulators and real-world data, hybrid methods augment
physics models with learned components. Actuator
networks~\cite{hwangbo2019learning,lee2020learning} capture complex motor dynamics,
while learned residuals~\cite{ajay2019combining} compensate for errors in simplified physics
models.
LIFT~\cite{huang2026towards} adopts a hybrid world model that combines
analytical rigid-body dynamics with learned residuals to support
off-policy optimization.
However, its model requires estimates of privileged
quantities.
Using a fully learned latent dynamics model,
SimDist~\cite{LevyJ-RSS-26} performs MPPI planning, introducing
additional online computational overhead at deployment.
Closer to our work are methods that replace the simulator with
a learned dynamics model and train policies directly inside it for real
hardware~\cite{li2025offline,li2025robotic}. 
In particular, uncertainty-aware offline MBRL~\cite{li2025offline} enables policy learning from fixed datasets without further environment interaction. However, our focus is a distinct deployment setting that of adapting an already pre-trained locomotion policy to a specific target dynamics that may lie outside its original training distribution. Rather than learning a policy solely from offline dataset, we investigate how simulation-acquired dynamics knowledge can be transferred to a new physical condition and used to improve a policy.

\noindent\textbf{Model Representations.} 
Latent world models such as
Dreamer~\cite{hafner2025mastering,wu2023daydreamer} compress observations into a
learned latent space and train policies on rollouts inside it, achieving strong
sample efficiency on raw sensory inputs. We take a different design in which the
dynamics model operates in an explicit physical state space, allowing privileged
simulation supervision to be injected through the teacher branch.
We further extend concurrent teacher-student learning~\cite{wang2024cts}
to both the policy and the dynamics model, jointly pre-training all modules
in a single stage instead of conventional two-stage
distillation~\cite{lee2020learning} for easy implementation. 
This design separates simulation-specific teacher
components from the proprioceptive student components used for subsequent
target-domain adaptation and deployment.

\section{Methodology}
\label{sec:method}

\subsection{Problem Formulation}
\label{sec:problem_formulation}

A domain adaptation problem can be formalized across two related Markov decision
processes (MDPs): a parameterized family of source-domain MDPs
$\mathcal{M}^s_\xi = (\mathcal{S}, \mathcal{A}, T^s_\xi, R, \gamma)$, indexed
by physical parameters $\xi \sim p(\xi)$ that span the domain randomization
distribution with transition $T^s_\xi$, and the target-domain MDP $\mathcal{M}^t = (\mathcal{S},
\mathcal{A}, T^t, R, \gamma)$ with transition $T^t$. The domains share the state space $\mathcal{S}$, action space
$\mathcal{A}$, task reward $R$, and discount factor
$\gamma \in (0,1)$.
The target dynamics $T^t$ may lie either within or outside the source family. 
Our objective is to improve policy performance under these specific target dynamics,
addressing either insufficient coverage by a narrow randomization distribution
$T^t \notin \{T^s_\xi : \xi \sim p(\xi)\}$, or 
potentially conservative behavior learned under a broad distribution
$\{T^t\} \subsetneq \mathcal{T}^s$ 
with large $\sup_{T_1,T_2\in\mathcal T^s}d(T_1,T_2)$.
A policy trained against a learned dynamics model $f_{\theta'}$
recovers performance in $\mathcal{M}^t$ only to the extent that $f_{\theta'}$
approximates the target dynamics. Let $\epsilon^t_m = \max_{s,a}
D_{\mathrm{TV}}\bigl(T^t(\cdot\,|\,s,a) \,\|\, f_{\theta'}(\cdot\,|\,s,a)\bigr)$
denote the model's worst-case error against $T^t$. Writing $\Delta T = T^t -
\bar{T}^s$, where $\bar{T}^s = \mathbb{E}_\xi[T^s_\xi]$ is the mean source
dynamics that is approximated by a pre-trained model $f_{\theta}$, the triangle
inequality gives
\begin{equation}
\epsilon^t_m \;\leq\; \underbrace{\|f_\theta - \bar{T}^s\|}_{\epre}
\;+\; \underbrace{\|f_{\theta'} - f_\theta - \Delta T\|}_{\eshift},
\label{eq:error_decomp}
\end{equation}
separating the target-domain model error into a pre-training term $\epre$ and
a shift term $\eshift$ that describes how well the parameter update
$\delta\theta = \theta' - \theta$ accounts for the dynamics difference between
source and target. This additional $\eshift$ term is absent in standard MBRL
and motivates our training architectural innovations. Our concurrent
teacher-student architecture preserves physical priors through the privileged
branch by capturing domain-invariant and hard-to-measure physics. The
proprioceptive branch then learns the target-specific deviation to minimize
$\eshift$.

Our approach operates in four main phases shown in Fig.~\ref{fig:framework}.
The first phase jointly trains the policy and the dynamics
model from simulator rollouts. The second phase collects real world data with diverse pre-trained behavior policies. Afterwards, the dynamics model is adapted to
the target domain with the batch of offline real-world trajectories. The model is
then used
as a surrogate environment for on-policy improvement, without
any further real-world interaction.

\begin{figure}[thb]
    \centering
    \includegraphics[width=0.48\textwidth]{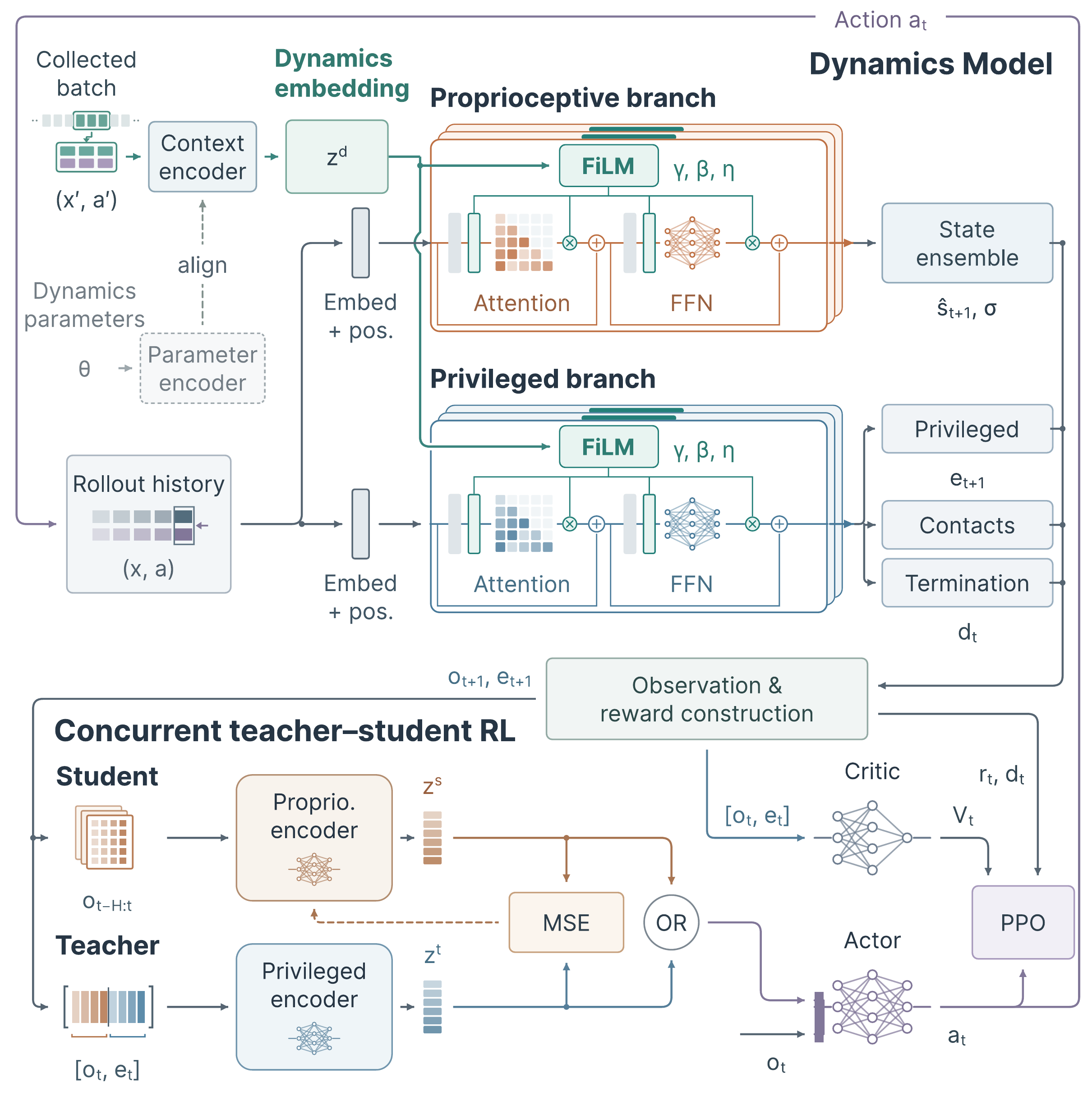}
    \caption{Overview of the proposed training architecture. The architecture
    consists of a concurrent teacher-student policy learning module (bottom) and dynamics
    modeling modules (top).
    }
    \label{fig:training_pipeline}
\end{figure}

\subsection{Online Pre-Training in Simulation}
\label{sec:pretraining}
The first phase in Fig~\ref{fig:framework}(a) fits both the policy and the dynamics model against the mean
source dynamics $\bar{T}^s$ using rollouts in a domain-randomized
simulator~\cite{NVIDIA_Isaac_Sim}, thereby minimizing the error $\epre$.
Extending a concurrent teacher-student framework~\cite{wang2024cts} to
dynamics modeling, this phase jointly trains four modules in a single stage: a
privileged teacher policy $\pi_t$, a proprioceptive student policy $\pi_s$, and
the corresponding privileged and proprioceptive branches of the dynamics model
$f_\theta$, as illustrated in Fig~\ref{fig:training_pipeline}.

\subsubsection{Observations}
At each time step $t$, the training observation is $s_t^p = [o_t, e_t]$,
where $o_t \in \mathbb{R}^{d_o}$ collects the robot's proprioceptive states,
and $e_t \in \mathbb{R}^{d_e}$ collects privileged simulation states that are
unavailable on real robot. The definition of these states is detailed in
Tab.~\ref{tab:observations}. The proprioceptive observation
history is $h_t = [o_{t-H}, \ldots, o_t]$ with length $H$.

\begin{table}[t]
    \centering
    \caption{Observation definitions}
    \label{tab:observations}
    \footnotesize
    \setlength{\tabcolsep}{1.5pt}
    \renewcommand{\arraystretch}{1.0}
    \begin{tabular*}{\columnwidth}
        {@{\extracolsep{\fill}}lcc@{\hspace{6pt}}lcc@{}}
        \toprule
        \multicolumn{3}{c}{\emph{Proprioceptive $o_t$}} &
        \multicolumn{3}{c}{\emph{Privileged $e_t$}} \\
        \cmidrule(lr){1-3}\cmidrule(lr){4-6}
        Quantity & Symbol & Dim. &
        Quantity & Symbol & Dim. \\
        \midrule
        Base angular vel. & $w$ & 3 &
        Base linear vel. & $v$ & 3 \\

        Projected gravity & $g$ & 3 &
        Joint torque/accel. & $\tau,\ddot{q}$ & 6 each \\

        Joint pos./vel. & $q,\dot{q}$ & 6 each &
        Foot vel./force & $v_f,F_f$ & 6 each \\

        Velocity commands & $v_c,w_c$ & 2, 1 &
        Relative base mass & $m$ & 1 \\

        Previous action & $a_{t-1}$ & 6 &
        Foot properties & $\phi$ & 6 \\

        Gait phase & $\mathrm{gait}_{\mathrm{pha}}$ & 2 &
        Base height & $h$ & 1\\

        Gait commands & $\mathrm{gait}_c$ & 3 &
        & & \\
        \bottomrule
    \end{tabular*}
\end{table}

\subsubsection{Teacher and Student Policies}
Teacher and student share one action network that acts on the proprioceptive
observation and a latent vector. The teacher policy $\pi_t(a_t \mid o_t, z^t)$,
where $z^t = E_p(s_t^p)$ is produced by a privileged encoder from the
privileged observation, is optimized with PPO to maximize the expected return.
Concurrently, the student policy $\pi_s(a_t \mid o_t, z^s)$, where $z^s =
E_o(h_t)$ is produced by a proprioceptive encoder from the observation history,
is distilled by matching the teacher's latent features, $L_{\text{distill}} =
\mathbb{E}\bigl[ \| z^s - \operatorname{sg}(z^t) \|^2 \bigr]$, with
$\operatorname{sg}(\cdot)$ the stop-gradient operator. Since $z^t$ encodes
privileged information, this distillation implicitly trains the student to
recover privileged-relevant features from the proprioceptive history $h_t$
alone
for deployment.

\subsubsection{Rewards}
The policy reward combines velocity tracking and foot-contact
objectives with regularization of posture, joint motion, effort,
and action changes.
The components and weights are listed in Tab.~\ref{tab:rewards}.

\begin{table}[t]
    \centering
    \caption{Training Rewards}
    \label{tab:rewards}
    \footnotesize
    \setlength{\tabcolsep}{0.7pt}
    \renewcommand{\arraystretch}{1.08}
    \begin{tabular*}{\columnwidth}{@{\extracolsep{\fill}}lcr@{}}
        \toprule
        Reward term & Value & Weight \\
        \midrule
        track linear velocity & $e^{-\frac{\|v_c-v_{xy}\|^2}{0.25\max(\|v_c\|,1)^2}}$ & $1.5$ \\
        track angular velocity & $e^{-\frac{(w_c-w_z)^2}{0.25}}$ & $0.75$ \\
        no fly & $\ind(\|F_{c}\|>0)$ & $1.0$ \\
        \midrule
        Penalty term & Value & Weight \\
        \midrule
        linear velocity penalty & $(v_z)^2$ & $-0.5$ \\
        angular velocity penalty & $(w_x)^2+(w_y)^2$ & $-0.05$ \\
        joint torques & $\tau^2$ & $-1.0\times10^{-5}$ \\
        joint acceleration & $\ddot q^2$ & $-2.5\times10^{-7}$ \\
        joint velocity & $\dot q^2$ & $-5.0\times10^{-5}$ \\
        joint power & $\sum_j|\tau_j\dot q_j|$ & $-2.0\times10^{-5}$ \\
        action rate & $(a_{t-1}-a_t)^2$ & $-0.01$ \\
        undesired contacts & $\sum_{i\in\mathcal B_{nc}}\ind(\|F_{c,i}\|>1)$ & $-0.25$ \\
        base height & $\|h-h_{\mathrm{ref}}\|^2$ & $-1.0$ \\
        flat orientation & $(g_x)^2+(g_y)^2$ & $-5.0$ \\
        swing force & $\sum_i(1-C_i^{\mathrm{des}})(1-e^{-\frac{\|F_i\|^2}{25}})$ & $-0.5$ \\
        stance velocity & $\sum_i C_i^{\mathrm{des}}(1-e^{-\frac{\|v_{f_i}\|^2}{0.25}})$ & $-0.5$ \\
        \bottomrule
    \end{tabular*}
    \par\vspace{3pt}
    \begin{minipage}{\columnwidth}\footnotesize
        $v_c$ and $w_c$ indicate user-commanded linear and angular velocity, respectively.
        $F_c$ is the total contact force vector summing all feet.
        $h_{\mathrm{ref}}=0.65\,\mathrm m$.
        $\mathcal B_{nc}$ contains the set of bodies excluding the feet. 
        $F_{c,i}$ is the
        contact force on the i-th body.
        $C_i^{\mathrm{des}}$ is the desired contact
        probability of foot $i$. $F_i$ and $v_{f_i}$ indicates the contact force and linear velocity
        of the i-th foot.
    \end{minipage}
\end{table}

\subsubsection{Dynamics Model}

Both branches of the dynamics model $f_\theta$ take histories of
$(x_t,a_t)$ as input, where $x_t=[w_t,g_t,q_t,\dot{q}_t]$
contains only the physical-state components of the
observation $o_t$.
A proprioceptive
branch predicts the next proprioceptive state $\hat{x}_{t+1}$, and a privileged
branch predicts the next privileged state $\hat{e}_{t+1}$ in parallel.
The proprioceptive branch is trained with a
multi-step autoregressive loss
\begin{equation}
  L_{\text{dyn}} = \frac{1}{K}\sum_{k=1}^{K} \| \hat{x}_{t+k} - x_{t+k} \|^2,
  \label{eq:dyn}
\end{equation}
where $\hat{x}_{t+k}$ is the model's $k$-step rollout from $(x_t,
a_{t:t+k-1})$ and $K$ is the prediction horizon. The privileged branch is
trained over the same horizon with teacher-forced inputs, 
and is
supervised with privileged targets that are unavailable at deployment. Both branches are conditioned on a dynamics embedding $z^d$
that characterizes the underlying dynamics, as detailed in
Section~\ref{sec:architecture}.

\subsection{Neural Dynamics Model Architecture}
\label{sec:architecture}

The proprioceptive and privileged branches share the same architectural
template using a causal Transformer backbone with an ensemble of multilayer
perceptron (MLP) prediction heads. The Transformer architecture follows the
GPT-2 design principles~\cite{radford2019language} to effectively capture
long-horizon dependencies and complex temporal dynamics. The ensemble provides
a per-step uncertainty signal to suppress policy exploitation of poorly
modeled regions.

\subsubsection{Input Representation and Tokenization}
The model operates on a sliding window of length $H_d=16$ containing the
proprioceptive observation history $x_{t-H_d:t}$ and applied action history
$a_{t-H_d:t}$. State and action tokens at each step are concatenated and
linearly projected into a $d_{\text{emb}}=384$-dimensional embedding space,
augmented with a learnable positional embedding to preserve temporal order.
The resulting sequence $X_t \in \mathbb{R}^{H_d \times d_{\text{emb}}}$ is fed
into the Transformer backbone.
A parameter encoder $E_{\mathrm{par}}$ maps simulation parameters
$\xi$ to a dynamics embedding $z^d=E_{\mathrm{par}}(\xi)$ and
is trained jointly with the dynamics model.
Subsequently, with both fixed, a context encoder
$E_{\mathrm{ctx}}$ learns to infer this embedding from
state--action windows $\mathcal{C}$ by minimizing
$
    L_{\mathrm{align}}
    = \mathbb{E}\!\left[
    \left\|E_{\mathrm{ctx}}(\mathcal{C})
    - \operatorname{sg}(z^d)\right\|_2^2
    \right].
    \label{eq:latent_align}
$
This enables dynamics conditioning from observed trajectories
when the physical parameters are unavailable.

\subsubsection{Causal Transformer Backbone}
The backbone is a decoder-only Transformer of $L=6$ stacked layers, each
consisting of a multi-head self-attention mechanism with $N_h=12$ heads and a
position-wise feed-forward network. A causal attention mask restricts each
token to attend only to past positions, ensuring that the prediction at time
$t$ depends only on history.
In both branches, $z^d$ conditions each attention and
feed-forward sublayer through feature-wise linear modulation
(FiLM). For sublayer $\ell$, normalized features are
modulated as
\begin{equation}
    \tilde{u}_{\ell}
    = \bigl[1+\gamma_{\ell}(z^d)\bigr]
    \odot \operatorname{LN}(u_{\ell})
    + \beta_{\ell}(z^d),
    \label{eq:film}
\end{equation}
where $\gamma_{\ell}$ and $\beta_{\ell}$ are learned affine
functions.
The sublayer's residual connection is additionally scaled
by $1+\eta_{\ell}(z^d)$, with a learned affine gate
$\eta_{\ell}$.

\subsubsection{Ensemble Heads}
The contextualized feature vector $h_t^{\text{last}} \in
\mathbb{R}^{d_{\text{emb}}}$ extracted from the final-position output is fed
into an ensemble of $N_e$ independent MLP heads $\{\phi_j\}_{j=1}^{N_e}$
sharing the same backbone, each predicting the next proprioceptive state as a
residual to the current state, $\hat{x}_{t+1}^{(j)} = x_t +
\phi_j(h_t^{\text{last}})$.
At training time
the heads are jointly optimized against the ground truth on bootstrap-resampled
mini-batches.
At rollout time their mean $\bar{x}_{t+1} =
\tfrac{1}{N_e}\sum_j \hat{x}_{t+1}^{(j)}$ serves as the state estimate, while
their variance $\Sigma_{t+1} = \mathrm{Var}_j(\hat{x}_{t+1}^{(j)})$ provides a
measure of epistemic uncertainty.

\subsubsection{Privileged Supervision Heads}
On privileged branch's backbone, three additional heads decode the
simulator's signals: (i) a regression head predicting the
privileged state $e_{t+1}$ under a mean squared error (MSE) loss; (ii) a
binary classification head predicting the contact state of each foot
under a binary cross-entropy (BCE) loss; and (iii) a
termination head predicting critical failure states under a BCE loss,
which stabilizes the value function during policy training. 

\subsection{Real-World Offline Fine-Tuning}
\label{sec:finetuning}
The third phase targets the shift term $\eshift$.
It adapts the dynamics model to the target domain using a batch of real-world
trajectories, then fine-tunes the simulator-pretrained policy on rollouts from
the adapted model. Notably, adapting the dynamics model requires real-world
data that covers the state--action regions the policy will explore, not merely
the trajectory from an optimal policy. Following the classical
system-identification intuition that accurate identification requires
excitation signals with broad frequency content, we collect $\mathcal{D}^t =
\{(x_{t_i}, a_{t_i}, x_{t_i+1})\}_{i=1}^{N_d}$ by deploying a heterogeneous pool
of behavioral policies on the physical robot, containing the
simulator-pretrained student policy, several near-converged checkpoints from
earlier training iterations, and variants with moderate action noise. By
contrast, data from a single policy covers only a narrow region of the
state--action space, leaving the dynamics model to hallucinate
plausible-but-wrong predictions outside this region. It is worth noting that
arbitrarily broad coverage is not necessary. Because PPO is on-policy and
constrained by a per-step trust region, the fine-tuned policy with limited update steps remains within a
bounded Kullback--Leibler (KL) neighborhood of the warm start, and only data within this neighborhood contributes to the
optimization. Our choice of behavioral pool is calibrated to span this
neighborhood without exceeding it, which also keeps real-robot data collection
safe because the deployed policies remain controllable rather than arbitrarily
aggressive.

The proprioceptive branch of the dynamics model is fine-tuned on
$\mathcal{D}^t$ using the same multi-step autoregressive loss
$L_{\text{dyn}}$ in~\eqref{eq:dyn}, while the privileged branch is frozen. 
Once the proprioceptive branch is aligned with the target, all dynamics-model
weights are frozen and the pre-trained policy $\pi_s$ is fine-tuned entirely
inside the model.
At each imagination
step, the proprioceptive branch predicts $\hat{x}_{t+1}$ with epistemic
uncertainty $\sigma_e$, and the privileged branch supplies the privileged
signals $\hat{e}_{t+1}$ that the task reward function depends on. The reward is
defined as
\begin{equation}
  r_t \;=\; r_{\text{task}}(\hat{x}_{t+1}, \hat{e}_{t+1}) \;+\; \beta \, \sigma_e,
  \label{eq:reward}
\end{equation}
where $r_{\text{task}}$ uses the same reward functions as during simulator
pre-training, $\sigma_e = \|\sqrt{\Sigma_{t+1}}\|_1$ collapses $\Sigma_{t+1}$
to a scalar by summing its per-dimension standard deviations,
and the penalty coefficient $\beta < 0$ discourages the policy
from visiting regions where the model is unreliable.
Rollouts are batched in parallel on the GPU,
requiring no further real-robot interaction during updates.

\section{Experimental Results}
\label{sec:results}

We evaluate our framework on a bipedal robotic platform in both simulation and
the real world, addressing the following research questions.
\textbf{(Q1)} Is the learned dynamics model a faithful enough surrogate to
support on-policy fine-tuning?
\textbf{(Q2)} Can our approach adapt the policy to extreme physical conditions
outside the training randomization range and outperform conservative
wide randomization range?
\textbf{(Q3)} How does our structure design contributes to the results?

The results can be visualized at
{\normalsize\urlstyle{same}\url{https://beyonddr.github.io/}}.

\subsection{Experimental Setup}
\label{sec:setup}

We evaluate our approach on a TRON~1 point-foot bipedal robot whose
self-weight is around 19~kg. The point-foot design minimizes the support
polygon, making the system underactuated and sensitive to mass and friction
perturbations. We test two extreme physical variations: (a) a heavy-payload
condition with 15~kg mounted at the torso, about 80\% of self-weight, and
(b) a low-friction condition on oil-lubricated ground. We compare our
approach against the following baselines.
(i) \emph{Narrow-DR}: a policy trained with nominal randomization (payload
$\in [-1, 3]$~kg, friction $\in [0.4, 1.2]$). Both target conditions lie
outside this range.
(ii) \emph{Wide-DR}: a policy trained with extensive randomization (payload
$\in [-1, 20]$~kg, friction $\in [0.05, 1.5]$) that technically covers both
targets.
(iii) \emph{RMA}~\cite{kumar2021rma}: implicit adaptation built on the Wide-DR
distribution.
The policy is
conditioned on a latent extrinsic vector inferred from the proprioceptive
history.
(iv) \emph{SysID}: a policy trained in simulation with the target physical
parameters identified from the collected data. The added mass and the friction coefficient are fitted with CMA-ES~\cite{hansen2001completely} by replaying recorded states and actions, similar to~\cite{tiboni2023dropo}. 
In our method, we retain $N=320$ intermediate checkpoints
during simulation pre-training as a candidate behavior-policy pool.
We further select 20 distinct policies from this pool and deploy them in the
target simulator for sim-to-sim experiments and on the physical robot for
sim-to-real experiments to collect around 30 min adaptation data.

\subsection{Model Fidelity Analysis}
\label{sec:model_fidelity}

\begin{figure}[!t]
    \centering
    \includegraphics[width=0.88\linewidth]{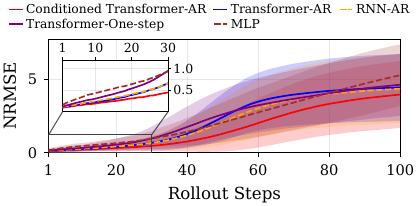}
    \caption{Normalized root-mean-squared error (NRMSE)
    of predicted states over a 100-step rollout across dynamics model
    architectures. Shaded bands denote one standard deviation across the
    evaluated trajectory windows. The inset enlarges the first 30 steps.}
    \label{fig:model_fidelity}
\end{figure}

A model-based framework only works if the dynamics model is a faithful enough
surrogate. We analyze the model along three axes: prediction accuracy across
architectures, hallucination behavior under different data-collection
strategies, and the fidelity of the privileged signals. To characterize compounding error through long horizon rollout, we
compare five models using the same 16-step history window:
(i) a vanilla MLP over the flattened history, trained with a one-step loss
(MLP); (ii) an RNN trained with a multi-step autoregressive loss
(RNN-AR)~\cite{li2025robotic}; (iii) a causal Transformer trained with a
one-step loss (Transformer-One-step); (iv) the same Transformer trained with
a multi-step autoregressive loss (Transformer-AR); and (v) our conditioned
Transformer-AR, which additionally modulates the backbone using
a dynamics embedding. The Transformer variants isolate the effects of
autoregressive training and explicit dynamics conditioning. During evaluation,
all models recursively feed back their predicted states.
We evaluate each model on 1000 randomly sampled trajectory windows drawn from
the simulator data with Narrow-DR and Wide-DR policy, each consisting of the
16-step conditioning history followed by 100 predicted steps, using the
normalized root-mean-squared error of the predicted proprioceptive
states at rollout step $k$:
\begin{equation}
    \operatorname{NRMSE}(k) = \frac{1}{M}\sum_{i=1}^{M}
    \sqrt{\frac{1}{d_o}\sum_{j=1}^{d_o}
    \left(\frac{\hat{x}_{i,k,j}-x_{i,k,j}}{\sigma_j+\varepsilon}\right)^2},
    \label{eq:rollout_nrmse}
\end{equation}
where $M=1000$ is the number of evaluated windows, $d_o$ is the predicted
state dimension, and $\hat{x}_{i,k,j}$ and $x_{i,k,j}$ are the predicted and
ground-truth values of state component $j$ at step $k$ in window $i$.
Here, $\sigma_j$ is the running standard deviation of component $j$ estimated
from the training data, and $\varepsilon=10^{-8}$ ensures numerical stability.
Fig.~\ref{fig:model_fidelity} shows that all baselines suffer increasing error
as $k$ grows, while our conditioned causal Transformer with autoregressive
consistently achieves the lowest error, confirming that the model design
minimizes compounding error over long horizons.

\begin{figure}[!t]
    \centering
    \begin{minipage}[c]{0.60\linewidth}
        \includegraphics[width=\linewidth]{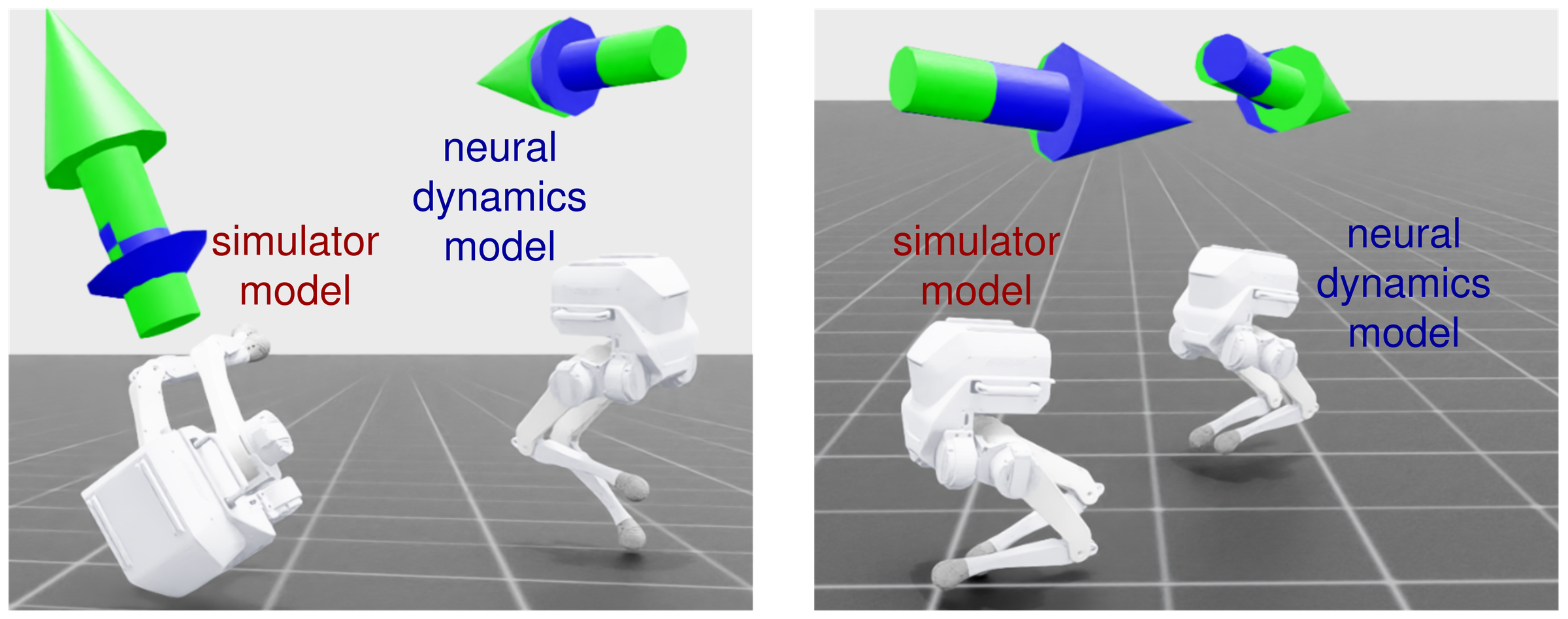}
    \end{minipage}\hfill
    \begin{minipage}[c]{0.375\linewidth}
        \includegraphics[width=\linewidth]{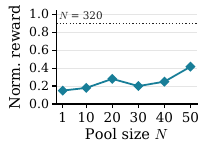}
    \end{minipage}
    \caption{Model hallucination. Left: the neural
    dynamics model (blue) predicts stable locomotion for a policy that fails
    catastrophically in the ground-truth simulator (red). Middle: training on
    data collected from diverse policies produces predictions that closely
    match the ground truth. Right: Policy reward versus behavioral
    pool size $N$ in held-out IsaacSim episodes.}
    \label{fig:model_hallucination}
\end{figure}

A separate concern is model hallucination. The learned dynamics model can
erroneously predict stable, high-reward trajectories in regions poorly covered
by the training data, even when its accuracy on the training distribution is
high. The left panel of Fig.~\ref{fig:model_hallucination} illustrates this. A
model trained on data from a single converged policy predicts that a
candidate policy achieves stable walking with consistent linear velocity,
while the same policy executed in the high-fidelity simulator collapses
immediately. 
Sampling from the behavioral policy pool yields a dataset that
retains operational stability while spanning a broader portion of the
state--action space, reducing the regions where the model hallucinates. The
middle panel of Fig.~\ref{fig:model_hallucination} shows that the resulting model's
predictions track the simulator ground truth closely.
To investigate the effect of behavioral diversity,
we vary the behavior-pool size from source domain while fixing the total collection budget
at 40\,000 transitions and evaluate the corresponding fine-tuned policies on held-out
nominal simulation episodes. As shown in the rightmost panel of
Fig.~\ref{fig:model_hallucination}, increasing the pool size from $N=1$
to $N=50$ raises the normalized return from 0.15 to 0.42.
The full configuration with $N=320$ achieves a
return of 0.90, suggesting that behavioral diversity helps mitigate model
hallucination and improves policy fine-tuning.

\begin{figure}[!t]
    \centering
    \includegraphics[width=0.95\linewidth]{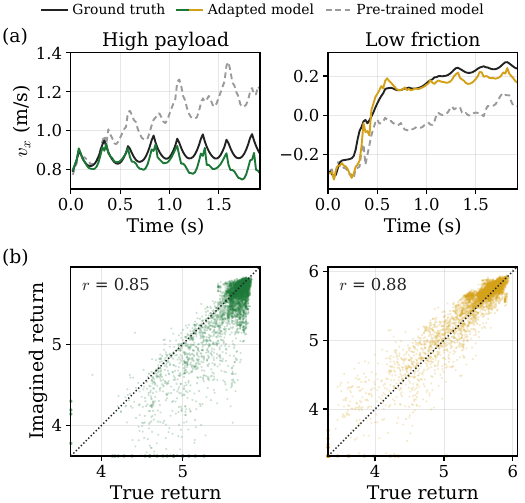}
    \caption{Accuracy of the predicted base velocity and imagined return. (a) Forward
  base velocity over one episode. (b) Imagined return
  versus the true return by the adapted model. Each point indicates the return value over one episode rollout from an adapted policy.}
    \label{fig:priv_fidelity}
\end{figure}

To further assess whether the dynamics model provides reliable reward
feedback for policy fine-tuning, we evaluate both privileged-signal
prediction and the fidelity of the resulting imagined returns.
Imagined rewards depend on privileged quantities, such as base
velocity, predicted by the privileged branch.
Figure~\ref{fig:priv_fidelity}(a) compares the pre-trained and
adapted models on a held-out episode.
Adaptation reduces the forward-velocity RMSE from 0.49 to
0.11~m/s under payload and from 0.15 to 0.085~m/s under low friction.
Figure~\ref{fig:priv_fidelity}(b) compares imagined and ground-truth
task returns, with each point representing the cumulative returns
over one episode generated by an adapted policy.
The returns are strongly correlated, with Pearson coefficients
of $r=0.85$ and $r=0.88$ under payload and low friction, respectively.
Together, these results support the use of the privileged branch
to construct informative reward signals for policy fine-tuning
within the dynamics model.

\subsection{Sim-to-Sim Transfer}
\label{sec:sim_eval}

\begin{figure}[!t]
    \centering
    \includegraphics[width=0.95\linewidth]{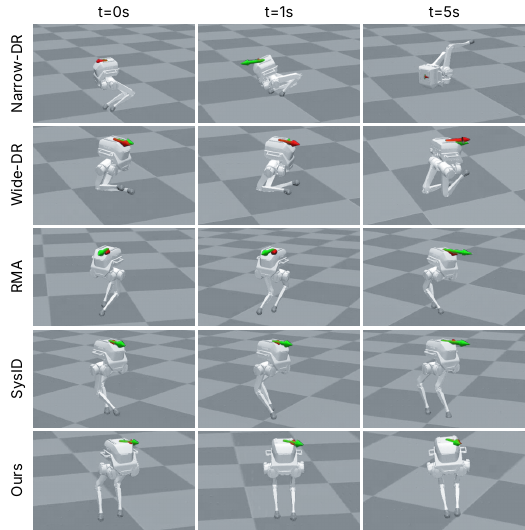}
    \caption{Qualitative comparison of the locomotion task under a 15~kg heavy
    payload in MuJoCo. Time-lapse snapshots illustrate the robot's behavior
    across the different baselines and our method.}
    \label{fig:mujoco_snap}
\end{figure}

\begin{table}[!t]
    \caption{Quantitative Performance in MuJoCo Under the High-Payload and
    Low-Friction Conditions}
    \label{tab:sim_metrics}
    \centering
    \fontsize{7.5}{9}\selectfont
    \newcommand{\simstat}[2]{#1$\pm$#2}
    \setlength{\tabcolsep}{1pt}
    \begin{tabular}{@{}lccccc@{}}
    \toprule
    Metric & Narrow-DR & Wide-DR & RMA & SysID & \textbf{Ours} \\
    \midrule
    \multicolumn{6}{@{}l}{\emph{High payload}} \\
    Succ.$\uparrow$ & \simstat{3.33}{2.31} & {\simstat{0.00}{0.00}} & \simstat{70.00}{25.87} & \simstat{96.00}{6.08} & \textbf{\simstat{97.33}{1.15}} \\[1pt]
    $E_v\downarrow$ & \simstat{1.29}{0.03} & - & \simstat{0.44}{0.17} & \simstat{0.32}{0.06} & \textbf{\simstat{0.24}{0.04}} \\[1pt]
    $E_w\downarrow$ & \simstat{0.89}{0.12} & - & \simstat{0.30}{0.16} & \simstat{0.18}{0.05} & \textbf{\simstat{0.15}{0.01}} \\[1pt]
    $\tau\downarrow$ & \simstat{31.26}{0.76} & - & \simstat{30.37}{1.89} & \simstat{32.09}{0.13} & \textbf{\simstat{28.22}{1.08}} \\[1pt]
    Pow.$\downarrow$ & \simstat{394.09}{37.13} & - & \simstat{133.68}{33.17} & \simstat{118.33}{16.76} & \textbf{\simstat{88.47}{5.32}} \\[1pt]
    $\ddot{q}\downarrow$ & \simstat{418.93}{59.21} & - & \simstat{144.10}{42.53} & \simstat{130.76}{19.43} & \textbf{\simstat{102.52}{6.05}} \\[1pt]
    \midrule
    \multicolumn{6}{@{}l}{\emph{Low friction}} \\
    Succ.$\uparrow$ & \simstat{53.67}{13.05} & {\simstat{0.00}{0.00}} & \simstat{94.67}{5.13} & \simstat{0.00}{0.00} & \textbf{\simstat{98.67}{1.15}} \\[1pt]
    $E_v\downarrow$ & \simstat{0.52}{0.09} & - & \simstat{0.28}{0.04} & \simstat{1.07}{0.03} & \textbf{\simstat{0.27}{0.04}} \\[1pt]
    $E_w\downarrow$ & \simstat{0.52}{0.13} & - & \simstat{0.17}{0.03} & \simstat{1.77}{0.25} & \textbf{\simstat{0.15}{0.02}} \\[1pt]
    $\tau\downarrow$ & \simstat{20.55}{0.69} & - & \textbf{\simstat{15.47}{0.49}} & \simstat{22.25}{1.09} & \simstat{16.55}{0.32} \\[1pt]
    Pow.$\downarrow$ & \simstat{206.56}{47.21} & - & \simstat{64.42}{10.76} & \simstat{454.01}{84.19} & \textbf{\simstat{54.58}{7.24}} \\[1pt]
    $\ddot{q}\downarrow$ & \simstat{290.45}{55.14} & - & \simstat{131.78}{11.63} & \simstat{453.01}{28.36} & \textbf{\simstat{105.22}{8.94}} \\[1pt]
    \bottomrule
    \end{tabular}
    \par\smallskip
    \begin{minipage}{\linewidth}\footnotesize
    Succ.: success rate in \%; $E_v$, $E_w$: root-mean-squared error (RMSE) of
    linear and angular velocity tracking (m/s, rad/s); $\tau$: joint torque RMS
    (N$\cdot$m); Pow.: mean absolute joint power (W); $\ddot{q}$: joint
    acceleration RMS (rad/s$^2$).
    The best value per metric is in bold within each condition.
    \end{minipage}
\end{table}

We evaluate the five methods transferred from IsaacSim~\cite{NVIDIA_Isaac_Sim} to MuJoCo~\cite{todorov2012mujoco} under both
target conditions. The assessment follows a continuous command-switching
protocol over a total duration of $T=1000$~s. During this period, the target
velocity command $v_{\text{cmd}}=[v_x, v_y, w_z]$ is resampled uniformly
every 10~s, resulting in $N_{seg}=100$ distinct command execution segments. The
sampling ranges are $v_x \in [-1.0, 1.0]$~m/s, $v_y \in [-0.5, 0.5]$~m/s, and
$w_z \in [-1.0, 1.0]$~rad/s. The success rate metric is the fraction of segments completed without failure, $(N_{seg} - N_{\text{fail}})/N_{seg}$.

Fig.~\ref{fig:mujoco_snap} shows time-lapse snapshots under
the payload condition and Tab.~\ref{tab:sim_metrics} reports the
quantitative comparison on both conditions over 3 training seeds. 
Narrow-DR lacks the torque capacity to counteract the
gravitational moment of the 15~kg load and collapses backward shortly after
initialization.
Wide-DR adopts a deeply crouched gait,
walking on its knees to lower the center of mass and remain stable across its
wide training distribution. This behavior does not satisfy the requirement for upright
bipedal locomotion and is therefore classified as a failure.
We consequently omit the other quantitative metrics for this baseline. 
While RMA and explicit SysID successfully preserve
stability, both SysID and RMA tracks less accurately and less efficiently.
Both limitations stem from their inherent design. 
RMA is still limited with wide randomization range, such that the policy tends to be conservative.
SysID is constrained by the bias of identified parameters.
Our method has a 45.4\% and 25.0\% improvement of velocity tracking error 
and 33.8\% and 25.2\% improvement of power
compared to RMA and SysID, respectively.
The reason is that our method allows the policy to converge to a
target-specific solution without the failure in a small randomization domain
or conservative performance loss inherent in
a large randomization domain.

\begin{figure}[!t]
    \centering
    \includegraphics[width=0.95\linewidth]{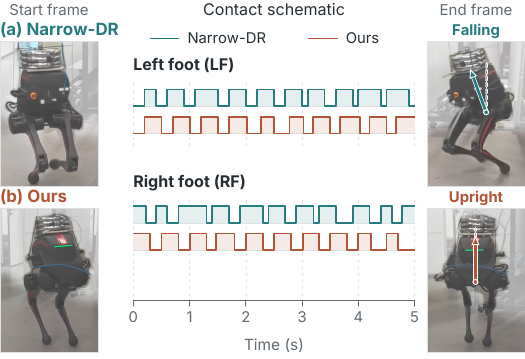}
    \caption{Real-world locomotion under a 15~kg payload. Snapshots and
    foot-contact sequences show that our method maintains a stable upright
    posture and highly periodic gait, while Narrow-DR fails to support the
    weight, resulting in severe body sagging and gait arrhythmia.}
    \label{fig:real_payload_snap}
\end{figure}

\begin{figure}[!t]
    \centering
    \includegraphics[width=0.46\textwidth]{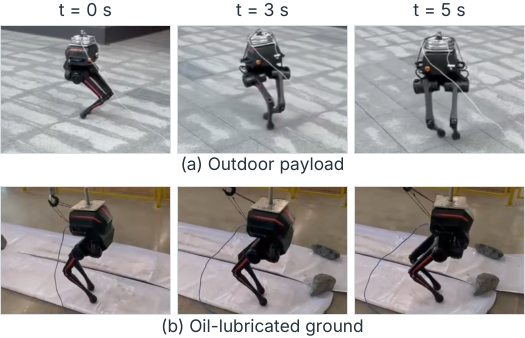}
    \caption{Real-world snapshots of the adapted policy.
(a) Outdoor walking with an added payload.
(b) Walking on oil-lubricated ground.}
    \label{fig:real_friction}
\end{figure}

Under the low-friction condition, the friction coefficient of the ground is reduced to 0.25. Narrow-DR adapts poorly, slipping and
falling in roughly half of the scenarios due to
aggressive ground reaction forces. Wide-DR similarly has a crouched gait.
RMA manages to maintain balance
but also adopts a more conservative posture with larger tracking error and power.
SysID fails with coefficient estimation 0.89 under the same recorded data as Ours. This biased value significantly degrades the performance.
Our approach successfully adapts to the slippery terrain, maintaining
precise tracking performance and energy efficiency.

\subsection{Ablation Study}
\label{sec:ablations}

\begin{table}[!t]
    \caption{Ablations of the Method's Design}
    \label{tab:ablations}
    \centering
    \footnotesize
    \setlength{\tabcolsep}{3pt}
    \begin{tabular*}{\columnwidth}{@{\extracolsep{\fill}}lcccc@{}}
    \toprule
    & \multicolumn{2}{c}{High payload}
    & \multicolumn{2}{c}{Low friction} \\
    \cmidrule(lr){2-3} \cmidrule(lr){4-5}
    Setting & Succ.$\uparrow$ & Score$\uparrow$
    & Succ.$\uparrow$ & Score$\uparrow$ \\
    \midrule
    Ours & 97.0 & 0.97 & 93.6 & 0.77 \\
    \midrule
    {w/o privileged branch} & 94.4 & 0.90 & 92.4 & 0.62 \\
    w/o teacher policy & 93.2 & 0.91 & 87.4 & 0.76 \\
    w/o dynamics embedding $z$ & 86.2 & 0.86 & 80.0 & 0.51 \\
    \midrule
    \multicolumn{5}{@{}l}{\emph{Uncertainty penalty $\beta$
    (Ours: $\beta=-1$)}} \\
    $\beta=0$  & 94.4 & 0.97 & 77.6 & 0.66 \\
    $\beta=-10$ & 83.4 & 0.74 & 60.4 & 0.50 \\
    \midrule
    \multicolumn{5}{@{}l}{\emph{Imagined rollout length (Ours: 100
    steps, 2~s)}} \\
    10 steps (0.2~s) & 91.2 & 0.93 & 86.6 & 0.76 \\
    1000 steps (20~s) & 71.4 & 0.67 & 12.6 & 0.15 \\
    \bottomrule
    \end{tabular*}
    \par\smallskip
    \begin{minipage}{\columnwidth}\footnotesize\raggedright 
    Evaluation on 500 trajectories in simulation for
    20~s.
    Succ.:
    success rate, in \%. Score: normalized return under the reward
    of Tab.~\ref{tab:rewards} 
    \end{minipage}
\end{table}

To assess the contributions of the model's each module, we conduct ablations through the privileged branch, the teacher policy and dynamics embedding condition of the training architecture's components in Fig~\ref{fig:training_pipeline}. We also examine the effects of the uncertainty regularization parameter $\beta$ in~\eqref{eq:reward} and imagined rollout length on policy fine-tuning.
Table~\ref{tab:ablations} reports success rates and normalized
returns over 500 simulated trajectories in IsaacSim per setting,
with a 20~s evaluation horizon and the reward defined in
Table~\ref{tab:rewards}.
The full method achieves the highest success rate and the
highest or tied-highest score under both conditions.
Among the structural components, dynamics conditioning provides
the largest benefit. Removing the dynamics embedding reduces
success rates by 10.8 and 13.6 percentage points under high payload
and low friction, respectively. The privileged branch and teacher
policy also contribute to performance, with their removal reducing
both success rates and returns.
A moderate uncertainty penalty with $\beta=-1$ improves low-friction
success by 16.0 percentage points over no penalty, whereas a stronger
penalty with $\beta=-10$ reduces performance under both conditions.
This shows that too small uncertainty penalty makes the policy to exploit uncertain region
and too large uncertainty penalty limits the policy's update due to less exploration.
Similarly, 100-step imagined rollouts outperform both shorter
and longer alternatives. With 1000-step rollouts, low-friction
success drops from 93.6\% to 12.6\%, consistent with the accumulation
of model errors. These results support using
moderate uncertainty regularization and a rollout horizon matched
to the model's predictive accuracy.

\subsection{Real-World Experiments}
\label{sec:real_world}

We conduct real-world experiments to evaluate whether our adaptation
framework enables a policy pre-trained under nominal DR to handle
challenging payload and low-friction conditions in the real world.
For each task, we collect around 30~min of real-world interaction data
using 20 distinct policies.
We use these data to adapt the dynamics model and subsequently
fine-tune the pre-trained policy through imagined interactions
with the adapted model.

Under the added 15kg payload, the nominal DR policy exhibits severe
body sagging and irregular stepping, whereas the adapted policy
maintains an upright posture and regular alternating foot contacts
throughout the deployment, as shown in
Fig.~\ref{fig:real_payload_snap}. Furthermore,
Fig.~\ref{fig:real_friction} shows the adapted policy
walking outdoors with payloads and on a low-friction surface
constructed by coating a plastic sheet with oil.
The robot maintains an upright posture and sustained locomotion
in both tasks.
We evaluate the adapted policy in five trials per target condition.
While the nominal DR policy fails in every trial under both
conditions, adaptation increases the success rate to 80\%
under the payload and 100\% 
on oil-lubricated ground. These results demonstrate the effectiveness of our method
in
challenging real-world dynamics.

\section{Conclusion}
\label{sec:conclusion}

We presented a domain adaptation framework that moves beyond static domain
randomization. Instead of expanding the training distribution to cover the
physical shift, the framework adapts a learned dynamics model to the
target condition using offline real-world trajectories, then re-optimizes the
policy entirely inside the aligned model via on-policy updates.
Experiments on a bipedal robot under a payload equivalent to 80\% of
its own weight and on oil-lubricated ground demonstrate the
effectiveness of this approach. These results point toward a broader
paradigm for sim-to-real transfer.
This paradigm has the potential to be
further improved as more real-world interaction
data become available.
Future work will focus on safe data collection, seeking to acquire informative real-world trajectories while minimizing risks to the robot. 










\bibliographystyle{IEEEtran}
\bibliography{references}

\end{document}